 \documentclass[preprint]{vgtc}                 

\graphicspath{{figures/}{pictures/}{images/}{./}} 

\usepackage{times}                     

\usepackage{amssymb}

\usepackage{tabu}                      
\usepackage{booktabs}                  
\usepackage{lipsum}                    
\usepackage{mwe}                       

\usepackage{algorithm}
\usepackage{algpseudocode}
\usepackage{amsmath}
\usepackage{multicol}

\usepackage{mathptmx}                  

\usepackage{amsmath}

\DeclareMathOperator*{\argmin}{arg\,min}

\onlineid{1079}

\vgtccategory{Short Paper}

\vgtcinsertpkg

\preprinttext{To appear at IEEE VIS 2024.}

\title{AEye: A Visualization Tool for Image Datasets}

\author{Florian Grötschla %
\and Luca A. Lanzendörfer %
\and Marco Calzavara %
\and Roger Wattenhofer   %
    }
    \affiliation{\scriptsize ETH Zurich, Switzerland \\ \{fgroetschla, lanzendoerfer, mcalzavara, wattenhofer\}@ethz.ch}

\teaser{
  \centering
  \includegraphics[width=\linewidth]{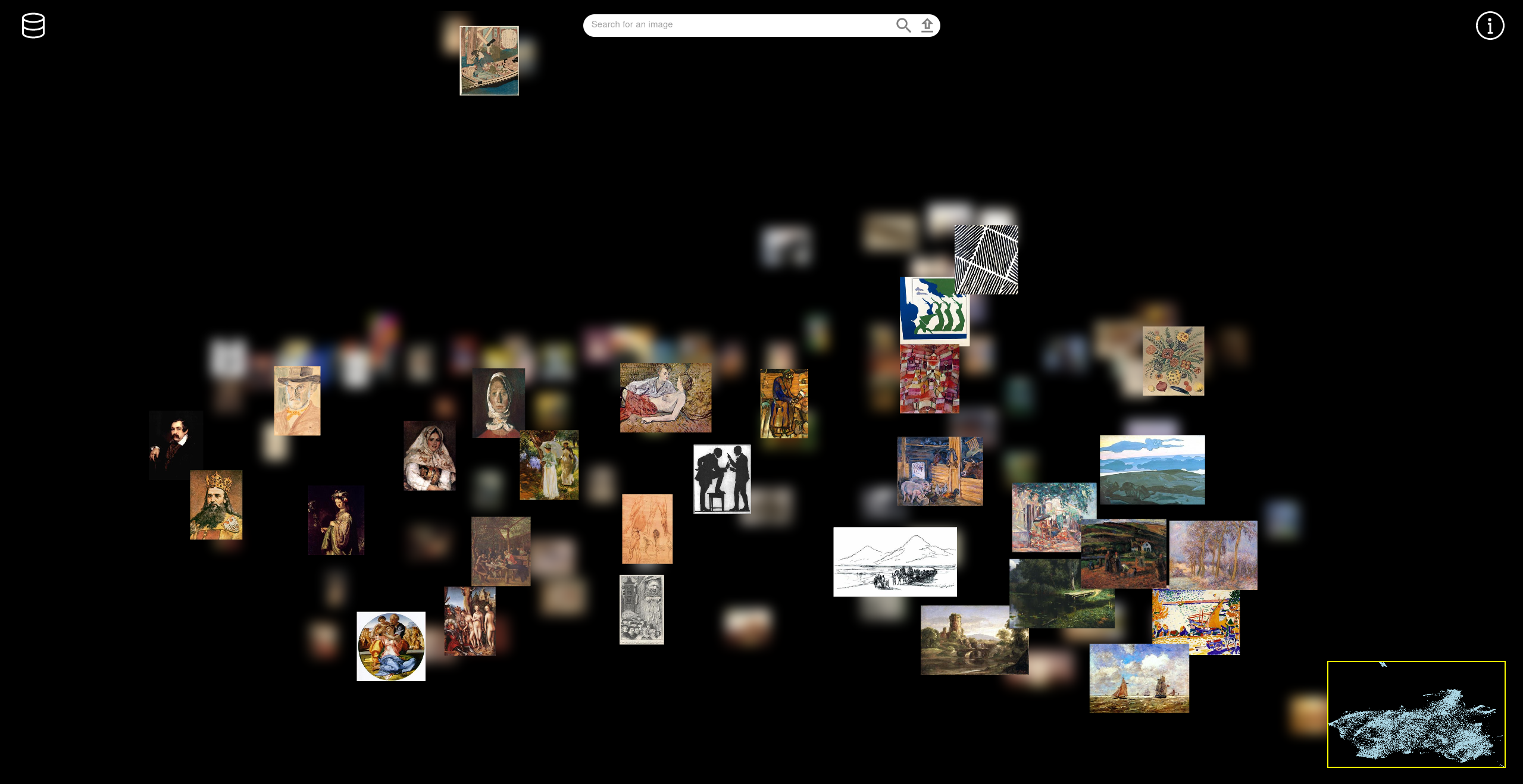}
  \caption{Overview of the AEye interface. Images are positioned according to their location in the CLIP embedding space and arranged in layers that the user can navigate by zooming. \textbf{Top left:} Dataset selector, \textbf{Top middle:} Search bar for semantic text and image search. \textbf{Top right:} Show information about the application. \textbf{Bottom right:} Minimap of the embedding space.}
  \label{fig:teaser}
}

\abstract{
    Image datasets serve as the foundation for machine learning models in computer vision, significantly influencing model capabilities, performance, and biases alongside architectural considerations. Therefore, understanding the composition and distribution of these datasets has become increasingly crucial. To address the need for intuitive exploration of these datasets, we propose AEye, an extensible and scalable visualization tool tailored to image datasets. AEye utilizes a contrastively trained model to embed images into semantically meaningful high-dimensional representations, facilitating data clustering and organization. To visualize the high-dimensional representations, we project them onto a two-dimensional plane and arrange images in layers so users can seamlessly navigate and explore them interactively. AEye facilitates semantic search functionalities for both text and image queries, enabling users to search for content. We open-source the codebase for AEye, and provide a simple configuration to add datasets. 
    
} 


\begin{document}


\firstsection{Introduction}

\maketitle

In today's data-driven landscape, the role of data in shaping the performance of artificial intelligence (AI) applications cannot be overstated. The quality, quantity, and complexity of the data significantly affect the performance and reliability of machine learning models across various domains.
As datasets continue to grow in size, researchers and practitioners face challenges in understanding and extracting meaningful insights, such as identifying patterns or outliers in the datasets. Traditional methods of data analysis often fall short when analyzing large-scale image datasets, highlighting the need for novel approaches to data exploration and visualization.

Effectively visualizing large-scale image datasets requires approaches that can distill visual information into semantically meaningful representations, enabling users to uncover patterns, trends, and anomalies within the data.
In response to these challenges, we introduce AEye -- a novel approach to visualizing image datasets. AEye leverages recent advancements in AI, specifically contrastive learning techniques, to embed semantic information into high-dimensional image representations. By projecting these representations onto a two-dimensional plane, AEye facilitates the visualization of image datasets in a manner that aligns with human perception and intuition.

We present the design and implementation of AEye and demonstrate its effectiveness in visualizing large image datasets. Through a series of demonstrative use cases, we illustrate how AEye enables researchers and practitioners to gain deeper insights into image data, uncover hidden patterns, and facilitate informed decision-making.
By providing AI-guided visualization, AEye offers a practical solution for visualizing large-scale image datasets and lets researchers and laymen extract insights from their data.

AEye is available at \url{aeye.ethz.ch}.

\begin{figure*}[h]
    \centering
        \includegraphics[trim=0 410 0 0,clip, width=\linewidth]{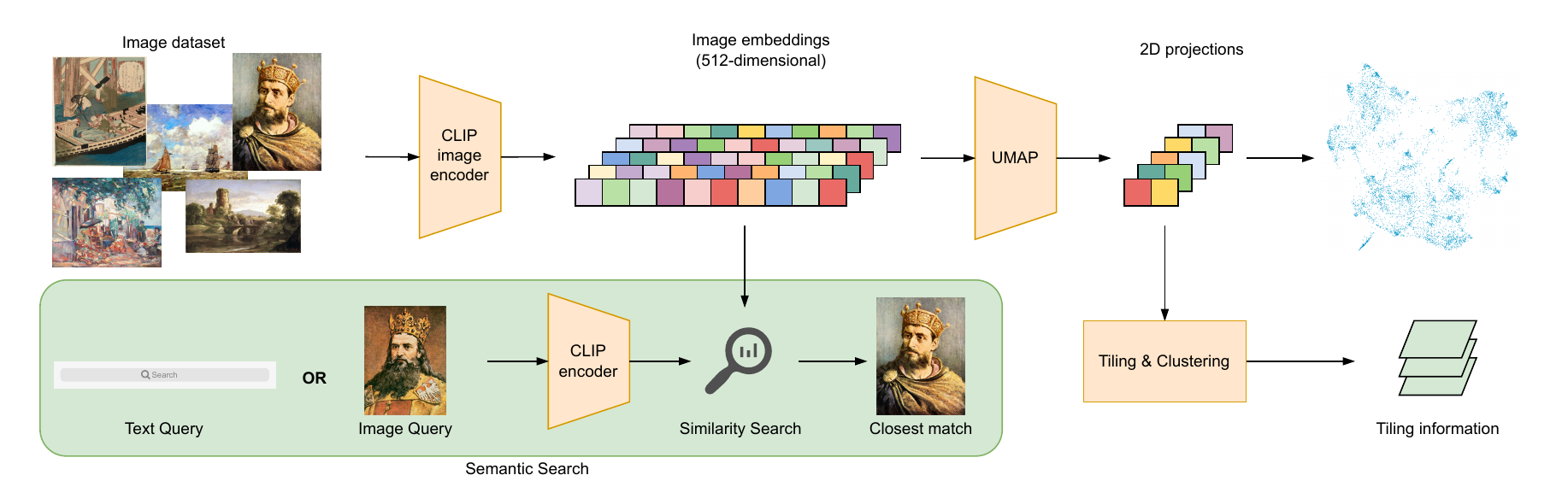}
    \caption{Architecture overview of AEye. Images are embedded with CLIP, stored in a vector database, and projected to a two-dimensional space with a UMAP projection (\Cref{sec:clip_and_umap}). The resulting positions are used for the visualization and a tiling and clustering module that computes representatives for each layer (\Cref{sec:clustering_and_tiling}). The semantic search takes text or image queries and uses CLIP to encode them. The vector database is used to find the nearest neighbor in the embedding space (\Cref{sec:search_and_captions}).}
    \label{fig:architecture}
\end{figure*}
\section{Related Work}
Large-scale data visualization is often facilitated by clustered and hierarchical representations~\cite{schultz2013open, schmidt2013vaico, klemm2014interactive}.
Dimensionality reduction techniques such as Principal Component Analysis (PCA)~\cite{wold1987principal}, t-Distributed Stochastic Neighbor Embedding (t-SNE)~\cite{van2008visualizing} or Uniform Manifold Approximation and Projection (UMAP)~\cite{mcinnes2018umap} have previously been applied for data visualization~\cite{10360935, afzal2023visualization}.
These techniques map high-dimensional data points to lower dimensions while clustering the data and making inherent patterns in the data more apparent. 
In our work, we choose UMAP for dimensionality reduction as it is more scalable than PCA and t-SNE while preserving local and global structures. 
Our work also touches upon the intuitive search for biases and imbalances in image datasets, which have previously been observed, and techniques were proposed to uncover and combat them~\cite{wang2022revise, fabbrizzi2022survey, Wang_2019_ICCV}. 

To obtain high-dimensional image embeddings, we use Contrastive Language-Image Pretraining (CLIP)~\cite{radford2021learning}, which learns to understand images and text simultaneously by embedding them in a shared latent space. 
CLIP is trained on a diverse range of image-text pairs from the internet, enabling it to learn robust and generalized representations that capture the semantic content of images across a wide spectrum of concepts and categories. 
CLIP embeddings capture rich semantic information for images and text, enabling a wide range of tasks such as image classification, image retrieval, and text-to-image generation~\cite{ramesh2021zero, che2023enhancing} without task-specific supervision. We use the pretrained OpenAI CLIP model to embed all images. 
CLIP and other contrastive learning-based approaches have been used for image visualization before~\cite{bohm2022unsupervised, 9984953}, mostly as point cloud visualizations.

The visualization technique most closely aligned to ours is the Embedding Projector~\cite{smilkov2016embedding}, which also visualizes embeddings generated by ML models with projected positions and offers a similar navigation technique consisting of zooming and panning. While it can also display images at the projected positions, it does not offer the layered visualization approach we provide. 

AEye builds on these previous works and adds novel methods to show only a representative selection of images with a layered visualization style. Using contrastive learning methods, we can ensure that embeddings maintain semantic information and facilitate additional search features. Lastly, our approach scales to larger datasets in terms of the visualization itself, which always remains comprehensible with not too many images on screen, and computational resources, which directly benefit from the former.

\section{AEye Application}
AEye is a web-based application designed to facilitate the exploration and comprehension of large-scale image datasets. At its core, AEye leverages the CLIP (Contrastive Language-Image Pretraining)~\cite{radford2021learning} embedding space to organize and visualize images in a two-dimensional plane. The positions of images within this embedding space are determined by their semantic similarity, allowing for intuitive navigation and exploration.

An overview of the AEye processing steps can be seen in \Cref{fig:architecture}.
In a data preprocessing stage, we compute CLIP embeddings for all images in the dataset, which are then stored in a vector database for fast nearest-neighbor lookups, which the semantic search relies on. 
The high-dimensional CLIP embeddings are then projected to two dimensions using the UMAP algorithm~\cite{mcinnes2018umap} to find spatial positions for all images. 
To accommodate the limited screen space and the large number of images in the dataset, AEye employs a layered visualization approach. Multiple layers are created that the user can navigate through. The last layer contains all images of the dataset at their projected positions, while the other layers only contain a selection of \emph{representatives}. As the user zooms in, the view transitions from layer to layer while zooming in on a continuously smaller area of the layers, which lets us populate them more and more densely while limiting the number of images on screen at any time. The representatives are chosen as the centers of a clustering we achieve with a modified k-means clustering algorithm~\cite{lloyd1982least}. This ensures that each layer provides a condensed yet informative dataset view. The process is outlined in more detail in ~\Cref{sec:clustering_and_tiling}. Lastly, we can compute AI-generated captions with LLaVA~\cite{liu2024visual} for all generated images in the preprocessing stage.

The visualization starts with a view of the first layer, which displays representative images from each cluster in the embedding space. As the user interacts with the visualization, they have the ability to zoom in on specific regions of interest, progressively revealing more detailed subsets of images from deeper layers. This interactive exploration enables users to uncover hidden patterns, clusters, and relationships within the dataset, empowering them to gain insights into the underlying structure of the data.

\subsection{CLIP Embeddings and UMap Projection}
\label{sec:clip_and_umap}

While CLIP embeddings effectively preserve the semantic meaning of images by encoding rich semantic information learned during pretraining~\cite{ramesh2021zero}, their high dimensionality, typically 512 dimensions, poses challenges for direct visualization. The sheer number of dimensions hinders intuitive interpretation and exploration of the embedding space. 
Therefore, we employ dimensionality reduction techniques to project them onto a two-dimensional plane to facilitate the visualization based on CLIP embeddings. This transformation enables us to represent the complex semantic relationships encoded in the embeddings in a more compact format. Among various dimensionality reduction techniques, we select UMAP due to its ability to preserve the data's local and global structure. Unlike traditional methods like PCA, which primarily focus on preserving variance, UMAP aims to capture the underlying manifold structure of the data, ensuring that nearby points in the high-dimensional space remain close together in the low-dimensional projection. 
An example of the projected embedding spaces can be seen on the minimap on the bottom right of~\Cref{fig:teaser}, the right side of~\Cref{fig:architecture}, as well as the bottom right of~\Cref{fig:mnist}. Projections like these are a common technique for the visualization of high-dimensional data~\cite{van2008visualizing, smilkov2016embedding} as the resulting point clouds usually exhibit a nicely clustered view of the embedding space. We use the projected embeddings as image positions throughout the application. As we cannot always show all images on screen, we further develop methods to select which images to show on the current layer through a clustering-based approach.

\subsection{Choosing Representative Images}
\label{sec:clustering_and_tiling}
\begin{figure}
    \centering
    \includegraphics[trim=50 40 170 490,clip,width=1.0\columnwidth]{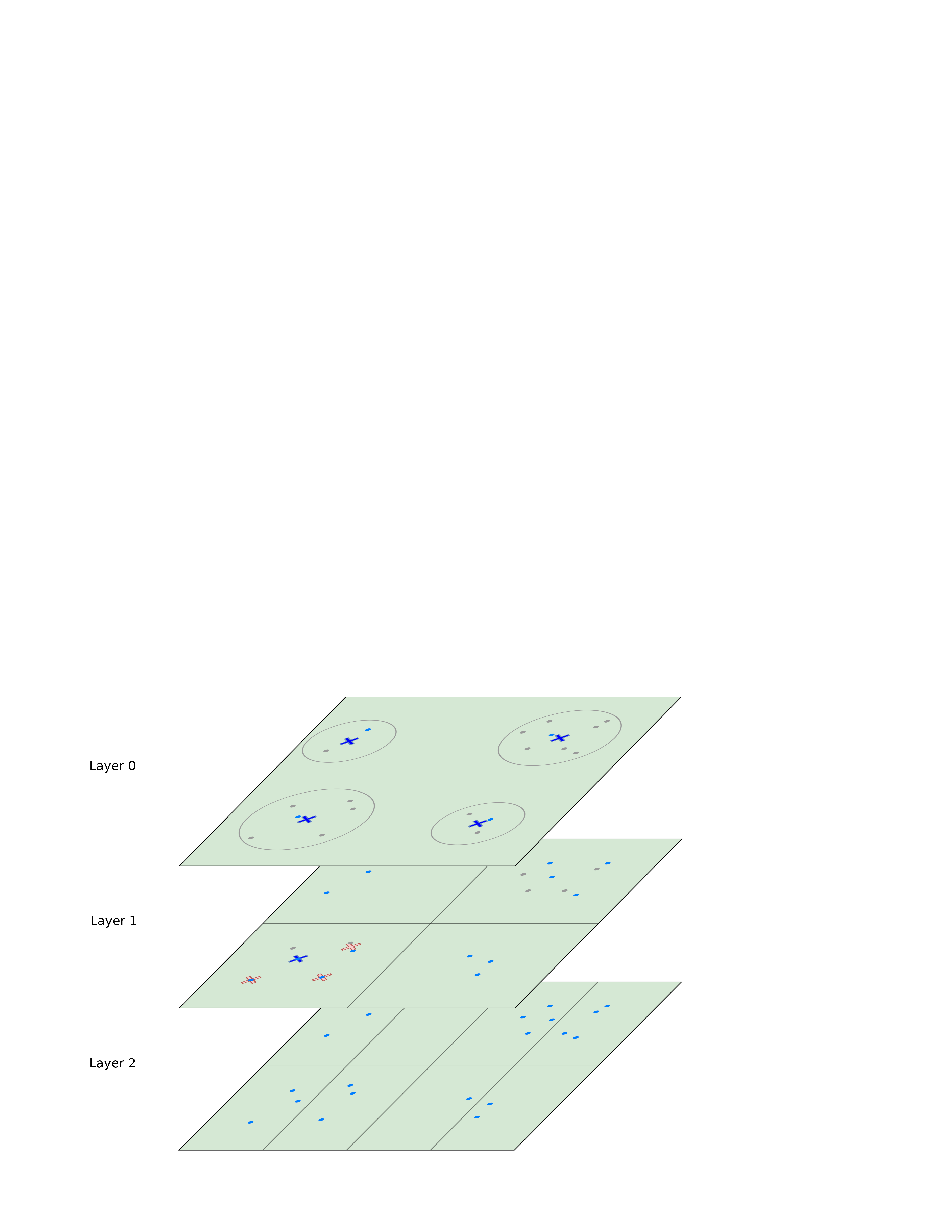}
    \caption{Visualization of the tiling hierarchy. Representatives for level 0 (blue dots) are obtained by clustering all points with k-means. Only the points closest to the centroids (crosses) are retained. In the next level, a k-means clustering is computed on every sub-tile, with the restriction of fixed centroids for the positions of representatives from the previous layers. Again, the closest points to the centroids are retained. This process finishes when all points can be kept for one level. Pseudocode can be found in~\Cref{alg:tiling}.}
    \label{fig:tiling}
\end{figure}
Large-scale image datasets contain an overwhelming number of images, necessitating a strategic approach to presenting a subset of images to users. We adopt a hierarchical strategy comprising multiple layers to address this challenge. In the initial layer, users encounter a limited set of representative images. By zooming in, users can traverse through these layers, progressively revealing additional images until the final layer displays all images. This hierarchical approach effectively manages the number of images on screen, resulting in a comprehensible visualization. The selection of representative images is guided by several criteria: they should provide a coarse-grained overview of the embedding space, maintain sufficient spacing between each other to avoid too many overlapping images, and reflect the characteristics of images within their respective area. Moreover, continuity in representation across layers is ensured by maintaining representatives from previous layers for all following ones. By doing so, images presented to the user in a previous layer do not disappear when zooming in but instead stay in place. We achieve these objectives through a clustering-based approach and tiling of the projected embedding space.

\Cref{fig:tiling} shows an overview of the proposed approach. Each layer consists of a regular grid of tiles, with the side length halving from one layer to the next. Within each tile, a fixed predefined number $k$ of images serves as representatives, approximately corresponding to the number of images visible on screen at any time. Selecting representatives is done by traversing layers from top to bottom and applying a k-means clustering algorithm with $k$ centers to each tile. Representatives are then chosen as the images closest to the cluster centers to ensure that they reflect the underlying structure of the embedding space. To maintain consistency in representatives across layers, we modify the k-mean algorithm by retaining the positions of representatives from previous layers as fixed centroids throughout the algorithm's execution. A detailed algorithm description in pseudocode is presented in~\Cref{alg:tiling}. When transitioning from layer to layer, the viewport's size is scaled proportionally with the size of the tiles, meaning that the viewport approximately covers the same number of tiles in every layer. As we limit the number of representative images per tile by $k$, the number of images on screen at the same time does not get too large.

\begin{algorithm}[t]
\small
\algrenewcommand\algorithmicindent{1.0em}%
\caption{Computation of Representatives}
\label{alg:tiling}
\begin{algorithmic}
\Require Image Dataset $I$, threshold $k$
\State $E(i) \gets \text{CLIP}(i) \quad \forall \ i \in I$
\State $\text{pos} : i \mapsto \mathbb{R}^2 \gets \text{UMAP}(E)$
\State $d \gets$ number of layers such that tiles on the last layer contain at most $k$ images.
\For{$l \gets 0$ to $d-1$}
\For{tile $t$ in $T_l$} \Comment{$T_l \sim$ all tiles in layer $l$}
\State $R_\text{prev} \gets \{ r \in R_{t'} \ | \ t' \in T_{l'}, l' < l\text{ and } r \in A(t) \}$ \Comment{$A(t) \sim$ area covered by tile $t$}
\State $\text{centers}_{\text{fixed}}\gets \{ \text{pos}(r) \ | \ r \in R_\text{prev}\}$
\State $I_t \gets \{ i \in I \ | \ \text{pos}(i) \in A(t)\}$ \Comment{images in current tile}
\State centers $\gets \text{k-mean}(I_t, \text{centers}_\text{fixed}, k)$ \Comment{k-mean with fixed centers}
\State $R_t \gets \{ \argmin_{i \in I_t}\left(\text{dist}(\text{pos}(i), c)\right) \ | \ c \in \text{centers} \}$
\EndFor
\EndFor
\end{algorithmic}
\end{algorithm}
This approach yields meaningful representatives, maintains scalability, and improves performance in the web application. While the computation of the k-mean algorithm poses a significant computational burden, especially in the first layer, subsequent layers benefit from its application to smaller subsets, albeit more numerous. These computations can be efficiently parallelized, and the number of required layers grows insignificantly with larger dataset sizes. Our demonstration webpage accommodates image datasets exceeding 100k images, where the preprocessing for the tiling and clustering took about as long as the generation of CLIP embeddings. This was on commodity hardware and took only few hours, even on the biggest datasets. As the number of layers is expected to scale logarithmically with the number of images, there will never be a need to transition through too many zoom layers.
Additionally, the generated tiling facilitates efficient data loading for the front end. As users navigate through layers, each tile consistently occupies a proportional screen space when in focus. These tiles serve as subdivisions, simplifying the selective loading of necessary data for visualization. The front end can request data for specific tiles, optimizing resource utilization and enhancing user experience.

\subsection{Semantic Search and AI Captions}
\label{sec:search_and_captions}
Beyond exploring visualized data through spatial navigation, AEye features semantic search functionality for text queries and images. Semantic search leverages the rich semantic representations encoded by the CLIP model for text and images, enabling users to retrieve relevant content easily, for example, with natural language.
For text queries, the semantic search first embeds the user-provided text query using the CLIP text encoder. Similarly, the CLIP image encoder embeds the user-provided image for image queries. 
To search for the nearest neighbor (with regard to cosine similarity), we store all image embeddings that we generate in the preprocessing stage in a vector database, more specifically, milvus~\cite{wang2021milvus}. 
Vector databases are specifically designed to handle high-dimensional vector data and offer fast and scalable search capabilities.
Therefore, when querying the database, we can use existing datastructures created in the preprocessing stage and reply quickly with the nearest neighbors sorted by similarity.

In addition to semantic search, AEye can also provide AI-generated image captions by running a captioning model on all images in the preprocessing phase. We use LLaVA~\cite{liu2024visual} for this purpose, as it resulted in the most accurate and descriptive captions in our testing, but other models can easily be integrated and used.
The captions provide valuable context and insights into the content of the images, enriching the user experience and providing feedback on how the AI model ``sees'' an image. Further, the captioning model is interchangeable, and the user can inspect the quality of captions for the used model.

\begin{figure}[t]
    \centering
    \includegraphics[width=\columnwidth]{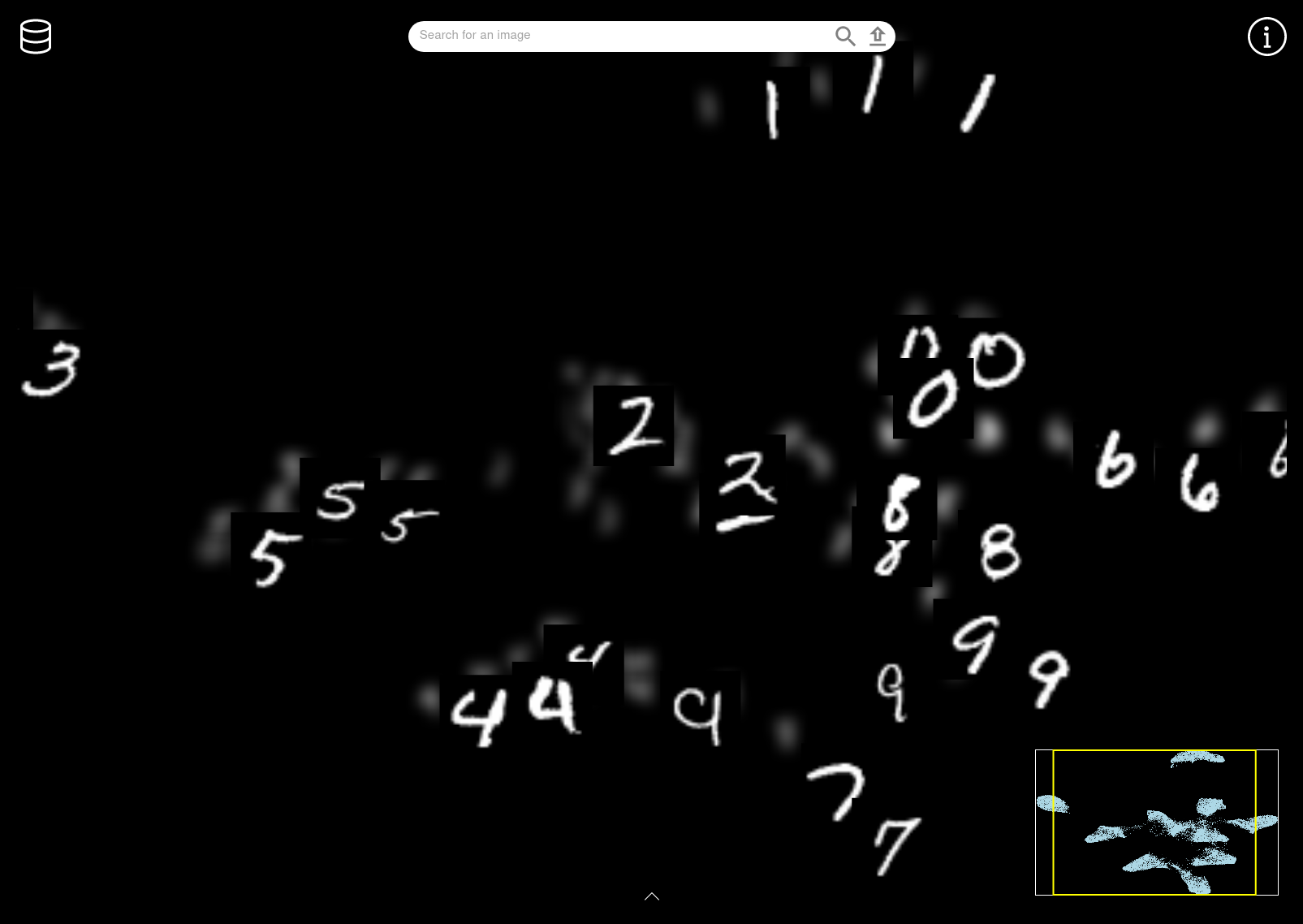}
    \caption{AEye view of the MNIST dataset. We observe that numbers are clearly separated by the projected CLIP embeddings, resulting in a meaningful clustering of the dataset. Similarly, the CelebA-HQ dataset shows a clear distinction between men and women.}
    \label{fig:mnist}
\end{figure}

\subsection{Interface Design}
\begin{figure}
    \centering
    \includegraphics[width=\columnwidth]{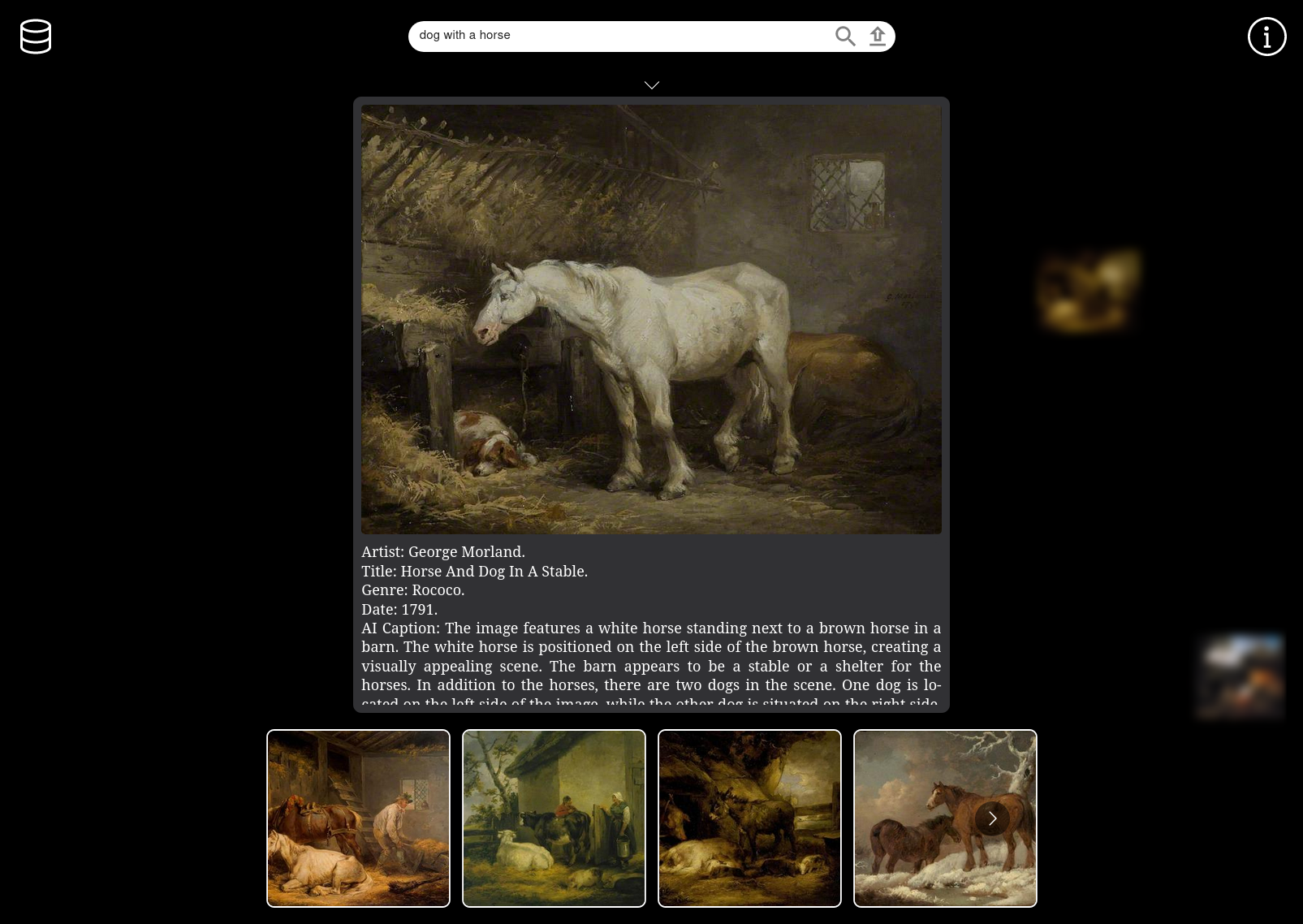}
    \caption{View of the application when searching for ``a dog with a horse.'' The nearest neighbors in the embedding space are presented below the search result. In addition to metadata provided by the dataset, an AI-generated caption of the image is shown.}
    \label{fig:enter-label}
\end{figure}

The user is initially presented with a few of the top layer of the embedding space and a search bar for the semantic text and image search in the center. If the user decides to submit a search query, the view zooms in to the closest match in the embedding space and shows the view depicted in~\Cref{fig:enter-label}, with an example for the MNIST dataset in~\Cref{fig:mnist}. 
For MNIST, we can clearly observe a semantic clustering of numbers in the projected embedding space, which lets us infer emerging patterns from the overview.

Users can further navigate the layered embedding space by zooming and moving the viewport around. Blurred previews are shown in the background to give a sense of the images in the next layer. Clicking on an image results in the same view as provided by the search, with more information provided by the dataset and an AI-generated caption of the image, as well as the closest neighbors in the CLIP embedding space. To stay oriented, a minimap with an overview of the whole embedding space is provided in the bottom right, as visible in~\Cref{fig:teaser}. Datasets can be selected on the top left, and an information button on the top right shows a small explanation for the application.

\subsection{Case Study}
To demonstrate a possible use case for AEye, we consider a hypothetical machine-learning practitioner working with the Common Objects in Context (COCO) 2017 dataset. The practitioner aims to improve their object detection model by first understanding the data. The COCO 2017 dataset contains 163,000 images, making it impractical to visualize all images at once or manually sift through them. This highlights the need for a comprehensive tool. The practitioner uses AEye to facilitate this process.

After the preprocessing, AEye produces its interactive visualization of the dataset on a 2D plane, clustering similar images together. This allows the practitioner to observe distinct clusters corresponding to different object categories, such as ``person,'' ``vehicle,'' and ``animal,'' revealing the distribution of categories and identifying under- or overrepresented ones. Several outliers were also detected, which, upon further examination, revealed labeling errors and unusual object combinations that could impact model performance.
The insights gained from AEye's visualization enable the practitioner to make informed decisions about the dataset. They identify underrepresented categories needing augmentation and corrected labeling anomalies, leading to a more balanced and accurate dataset for training their object detection model.

\section{Conclusion}
AEye offers a comprehensive solution for visualizing large-scale image datasets, leveraging contrastively trained embedding models for semantically rich representations. By incorporating hierarchical tiling, clustered subspaces, and semantic search features alongside AI-generated captions, AEye facilitates intuitive navigation and exploration of diverse image collections. The scalability and extensibility of AEye enable researchers and other users to explore various datasets, from machine learning to general art collections. 
With AEye, unlocking insights and uncovering patterns within large-scale image datasets becomes both accessible and insightful. A demonstration website with a selection of datasets,~\footnote{\href{aeye.ethz.ch}{aeye.ethz.ch}} as well as the source code, can be found online.~\footnote{\href{https://github.com/ETH-DISCO/aeye}{https://github.com/ETH-DISCO/aeye}}

\pagebreak

\bibliographystyle{abbrv-doi}

\bibliography{template}
\end{document}